\documentclass[letterpaper, 10 pt, conference]{ieeeconf}  

\IEEEoverridecommandlockouts                              

\usepackage{url}
\usepackage{amsmath}
\usepackage{graphicx}
\usepackage{verbatim}
\usepackage{algorithm}
\usepackage{algorithmic}
\usepackage{hyperref}
\usepackage{color}
\usepackage{wrapfig}
\usepackage[small]{caption}
\usepackage[table]{xcolor} 

\long\def\ignorethis#1{}

\newcommand{\nth}{\text{th}}
\newcommand{\pr}{^\prime}
\newcommand{\tr}{^\mathrm{T}}
\newcommand{\inv}{^{-1}}

\newcommand{\gauss}{\mathcal{N}}

\newcommand{\vnorm}[1]{\|#1\|}

\newcommand{\costnorm}{r_\ell}

\newcommand{\cost}{\ell}
\newcommand{\state}{\mathbf{x}}
\newcommand{\action}{\mathbf{u}}
\newcommand{\hstate}{\hat{\mathbf{x}}}
\newcommand{\haction}{\hat{\mathbf{u}}}

\newcommand{\traj}{\tau}

\newcommand{\detpolicy}{g}

\newcommand{\priordyn}{\tilde{p}}

\newcommand{\xu}{{\state\action}}
\newcommand{\fc}{f_c}
\newcommand{\fx}{f_\state}
\newcommand{\fu}{f_\action}
\newcommand{\fxu}{f_{\state\action}}
\newcommand{\fct}{f_{c t}}
\newcommand{\fxt}{f_{\state t}}
\newcommand{\fut}{f_{\action t}}

\newcommand{\fyt}{f_{\state\action t}}

\newcommand{\Qxt}{Q_{\state t}}
\newcommand{\Qut}{Q_{\action t}}
\newcommand{\Qyt}{Q_{\state\action t}}
\newcommand{\Qxxt}{Q_{\state,\state t}}
\newcommand{\Quut}{Q_{\action,\action t}}
\newcommand{\Quxt}{Q_{\action,\state t}}

\newcommand{\Qyyt}{Q_{\state\action,\state\action t}}

\newcommand{\Vxt}{V_{\state t}}
\newcommand{\Vxxt}{V_{\state,\state t}}
\newcommand{\Vxtp}{V_{\state t+1}}
\newcommand{\Vxxtp}{V_{\state,\state t+1}}
\newcommand{\kpol}{\mathbf{k}}
\newcommand{\Kpol}{\mathbf{K}}

\newcommand{\noise}{\mathbf{F}}

\newcommand{\costgradt}{\cost_{\state\action t}}
\newcommand{\costhesst}{\cost_{\state\action,\state\action t}}

\newcommand{\st}{\state_t}
\newcommand{\at}{\action_t}

\newcommand{\sth}{\hstate_t}
\newcommand{\ath}{\haction_t}

\newcommand{\empsig}{\hat{\Sigma}}
\newcommand{\empmu}{\hat{\mu}}
\newcommand{\empn}{N}
\newcommand{\priorphi}{\mathbf{\Phi}}
\newcommand{\priormu}{\mu_0}
\newcommand{\priormut}{\mu_{0t}}
\newcommand{\priorm}{m}
\newcommand{\priorn}{n_0}
\newcommand{\datapt}{\mathbf{p}}
\newcommand{\ddpdiscount}{\gamma}
\newcommand{\onlinediscount}{\beta}
\newcommand{\xxt}{\Delta}
\newcommand{\net}{\bar{f}}
\newcommand{\sigt}{\bar{\Sigma}_{\state\action,\state\action}}
\newcommand{\sigp}{\bar{\Sigma}_{\state\pr,\state\pr}}

\title{\LARGE \bf
One-Shot Learning of Manipulation Skills with Online Dynamics Adaptation and Neural Network Priors
}

\author{Justin Fu, Sergey Levine, Pieter Abbeel
\thanks{Department of Electrical Engineering and Computer Science, University of California, Berkeley, Berkeley, CA 94709}%
}

\begin{document}

\maketitle
\thispagestyle{empty}
\pagestyle{empty}

\begin{abstract}

One of the key challenges in applying reinforcement learning to complex robotic control tasks is the need to gather large amounts of experience in order to find an effective policy for the task at hand. Model-based reinforcement learning can achieve good sample efficiency, but requires the ability to learn a model of the dynamics that is good enough to learn an effective policy. In this work, we develop a model-based reinforcement learning algorithm that combines prior knowledge from previous tasks with online adaptation of the dynamics model. These two ingredients enable highly sample-efficient learning even in regimes where estimating the true dynamics is very difficult, since the online model adaptation allows the method to locally compensate for unmodeled variation in the dynamics. We encode the prior experience into a neural network dynamics model, adapt it online by progressively refitting a local linear model of the dynamics, and use model predictive control to plan under these dynamics. Our experimental results show that this approach can be used to solve a variety of complex robotic manipulation tasks in just a single attempt, using prior data from other manipulation behaviors.

\end{abstract}

\section{Introduction}

One of the remarkable features of human and animal motor control is the ability to quickly adapt to new situations. When a child is asked, for example, to stack two unfamiliar Lego blocks, he or she might play with the objects and take a small amount of time to learn about their dynamics, but will typically succeed at the task very quickly. This is sometimes referred to as one-shot learning, where an agent must successfully perform a task given one, or very few attempts. Reinforcement learning (RL) provides a computational framework for robots to learn new motor skills, but typical applications of RL focus more on mastery than one-shot learning, and require a substantial number of training episodes \cite{dnp-spsr-13}. Model-based RL methods reduce the required interaction time by acquiring a model of the system dynamics, and using this model to discover an effective policy. However, model-based RL algorithms that operate in this way must be equipped with a model that can represent a good approximation to the dynamics, and they must be provided with sufficient experience to optimize this model to produce accurate dynamics predictions \cite{kbp-rlrs-13}. A single attempt at the task is often insufficient to obtain such an accurate model, and while these challenges can be mitigated by incorporating domain knowledge about the system \cite{np-umkli-10,ch-erlri-15} or demonstrations \cite{wd-tosli-10}, they make the development of a general-purpose one-shot learning method difficult.

\begin{figure}
\setlength{\unitlength}{0.5\columnwidth}
\begin{picture}(1.99,1.5) \linethickness{0.5pt}

\includegraphics[width=240pt]{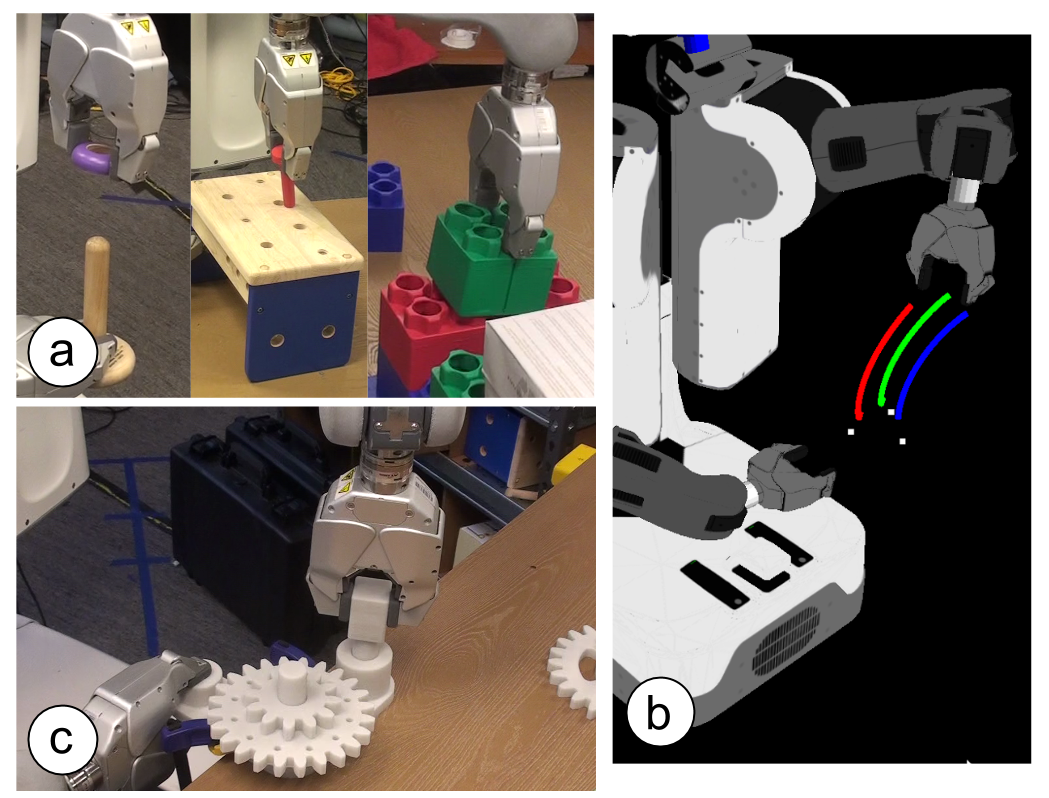}

\end{picture}
\caption{Diagram of our method: the robot uses prior experience from other tasks (a) to fit a neural network model of the dynamics of object interaction tasks (b). When faced with a new task (c), our algorithm learns a new model online during task execution, using the neural network as a prior. This new model is used to plan actions, allowing for one-shot learning of new skills.
\label{fig:teaser}
}
\vspace{-0.3in}
\end{figure}

We propose to address the first challenge by developing a method that can use a coarse model of the system dynamics that is adapted online to better match the most recent experience. This frees the method from needing a representation that can capture global dynamics with high accuracy, and requires only the ability to accurately represent the dynamics locally, which we show can be done with a simple linear model. The accuracy of the global model still affects the proficiency with which a new task can be performed, since more accurate models require less adaptation, but even an inaccurate initial model is sufficient for our method to perform a variety of manipulation tasks on the first attempt. We show that even a simple linear global model can allow this framework to perform a variety of manipulation tasks with nonlinear dynamics, while a more sophisticated nonlinear model based on neural networks greatly improves the success rate on more complex tasks. Online adaptation also offers us a way to address the second challenge: since the global model does not need to be very accurate, it can be estimated using data from tasks that are different from the one being attempted. This enables our approach to perform one-shot learning of new tasks by using the robot's prior experience on \emph{other} tasks, without requiring explicit domain knowledge or demonstrations to be provided by the designer. A diagram of our method is shown in Figure~\ref{fig:teaser}. We demonstrate our approach by learning a variety of challenging, contact-rich manipulation behaviors on a PR2 robot. All of our motion skills involve low-level torque control of the robot's motors, and high dimensional state, corresponding to joint angles, joint velocities, and end-effector pose.


\section{Related Work}
\label{sec:related}



Reinforcement learning techniques have been applied to a range of robotic control problems \cite{kbp-rlrs-13}. Such methods can be broadly categorized as model-free methods, which directly learn a control policy from system interaction \cite{ps-rlmsp-08}, and model-based methods, which first learn a model of the system dynamics, and then optimize the policy under this model \cite{dnp-spsr-13}. Model-based methods can be substantially more sample-efficient and typically achieve the fastest learning times, but they require a model representation that can be used to learn an accurate estimate of the true dynamics. Typically, in order to learn the model from a small amount of interaction, they make various assumptions, such as smoothness \cite{dr-pmbde-11,bswr-artoc-14} or access to prior knowledge about the system \cite{np-umkli-10,ch-erlri-15}, since the general problem of learning a complex, nonlinear function from a small amount of data is difficult. Contact-rich robotic manipulation skills present an especially challenging model learning problem, and simple assumptions, such as smoothness and prior knowledge, are often insufficient to acquire a model that is accurate enough for manipulation.

In this paper, we instead specifically seek to learn new behaviors with a coarse model of the system dynamics that does not adequately capture all of the intricacies of the real system, and locally adapt this model online based on the most recent experience. In this way, our approach resembles adaptive control \cite{aw-ac-94}. However, unlike standard adaptive control, we incorporate an expressive (but potentially inaccurate) prior model of the dynamics, constructed from previous experience on other manipulation tasks, and use only high-level objectives, such as the desired position of a target object, rather than simple trajectory tracking. This makes our method suitable for complex robotic manipulation skills with only high-level specification in the form of a cost function. Experience-based priors have previously been suggested in reinforcement learning \cite{wgrkt-bayesprior-11}, though typically in the context of accelerating iterative model-free learning. In contrast, our work demonstrates one-shot learning, where the robot can immediately perform new tasks by leveraging prior experience.

In order to choose the actions under our locally adapted model, we use model predictive control (MPC) based on differential dynamic programming (DDP) \cite{tet-mjc-12}. Neural network models have recently been combined with MPC for planar task-space control of cutting tasks \cite{lks-dmpc-15}. While this approach is similar in spirit to the one presented in this work, we tackle a wider range of robotic manipulation tasks using direct, low-level torque control of the entire 7 degree of freedom arm. Since the tasks that we tackle are quite different from the prior experience used to train the neural network, local adaptation is required for success, as shown in our experimental results. Outside of robotic manipulation, MPC has been combined with online and offline adaptation \cite{fks-ampcc-07,abt-elbmp-12,cmhh-clam-13}, but typically in the context of trajectory tracking, rather than learning new skills with high-level specification.


\section{Background}

In reinforcement learning, as well as in optimal control, the aim is to control a dynamical system given by $\state_{t+1} = f(\st,\at)$ by choosing the actions $\at$ to minimize the total cost $\sum_{t=1}^T \cost(\st,\at)$, where $T$ is the time horizon. We consider fixed-horizon episodic tasks in this paper, though our method can also be extended to an infinite horizon formulation. When the system dynamics $f(\st,\at)$ are unknown, we can construct an estimate of the dynamics $\hat{f}(\st,\at)$, for example by fitting a parameterized function approximator to data from prior system interactions, and then optimize a sequence of actions under the estimated dynamics $\hat{f}(\st,\at)$.

We will make use of the iterative linear quadratic regulator (iLQR) algorithm \cite{lt-ilqr-04} to optimize the actions with respect to the cost under an estimated dynamics model. This algorithm can be viewed as an application of the Gauss-Newton method for trajectory optimization, and requires iteratively linearizing the dynamics around the current nominal trajectory, denoted $\hat{\traj} = \{\hat{\state}_1,\hat{\action}_1,\dots,\hat{\state}_T,\hat{\action}_T\}$, constructing a quadratic approximation to the cost, computing the optimal actions with respect to this approximation of the dynamics and cost, and running the resulting actions forward to obtain a new nominal trajectory. In the derivation below, we will assume that the linearized stochastic dynamics are given by $p(\state_{t+1}|\st,\at) = \gauss(\fxt\st + \fut\at + \fct, \noise_t)$, where $\noise_t$ is the covariance of the Gaussian dynamics noise, and the quadratic approximation to the cost consists of a linear term $\costgradt$ and a quadratic term $\costhesst$, where the subscripts denote differentiation with respect to the vector $[\st; \at]$. When the dynamics are linear and the cost is quadratic, the Q-function and the value function are both quadratic, and given by
\begin{align*}
V(\st) &= \frac{1}{2}\st\tr\Vxxt\st + \st\tr\Vxt + \text{const} \\
Q(\st,\at) &= \frac{1}{2}[\st;\! \at]\tr\Qyyt[\st;\! \at] \!+\! [\st;\! \at]\tr \Qyt \!+\! \text{const}
\end{align*}
We can express them with the following recurrence:
\begin{align*}
\Qyyt &= \costhesst + \ddpdiscount\fyt\tr\Vxxtp\fyt \\
\Qyt &= \costgradt + \ddpdiscount\fyt\tr\Vxtp \\
\Vxxt &= \Qxxt - \Quxt\tr\Quut\inv\Quxt \\
\Vxt &= \Qxt - \Quxt\tr\Quut\inv\Qut,
\end{align*}
\noindent which allows us to compute the optimal control law as \mbox{$\detpolicy(\state_t) = \haction_t + \kpol_t + \Kpol_t(\state_t - \hstate_t)$}, where $\Kpol_t = -\Quut\inv \Quxt$ and $\kpol_t = -\Quut\inv \Qut$. Performing a forward rollout using this control law allows us to find a new nominal trajectory, and the backward dynamic programming pass is repeated around this trajectory to take the next Gauss-Newton step. The discount factor $\ddpdiscount$ allows us to reduce the weight on later states. While DDP is typically used with $\ddpdiscount = 1$, we found that we could obtain better results under uncertain, estimated dynamics with $\ddpdiscount = 0.95$. Intuitively, this reflects the fact that predictions further into the future are less likely to correspond to reality under uncertain dynamics, so it makes sense not to weight them as highly.

Iterative LQR can be used to optimize a trajectory offline under known system dynamics, or even with dynamics estimated from previous executions on a physical system \cite{la-lnnpg-14}. However, in this work we instead use iterative LQR to perform model-predictive control, where the policy is recomputed in real time at each time step to update the next action in response to the current state. Iterative LQR is well suited for MPC because it is fast, can be effectively used with short horizons, and readily allows warm-starting with the previous solution \cite{tet-sscbo-12}.

However, using iterative LQR for MPC still requires us to obtain an estimate of the linearized system dynamics, given by $\fxt$ and $\fut$. The dynamics can be estimated from a simulator of the physical system, but this requires considerable knowledge about the system being controlled. In the case of robotic manipulation, it is reasonable to expect to obtain a good model of the robot, but we often lack a model of the objects that the robot is interacting with, so being able to handle unknown dynamics is highly desirable. In Section~\ref{sec:related}, we discussed how previous methods have approached this problem by using a variety of physical and statistical estimates. In this work, we take a different approach, and do not attempt to accurately learn a global model of the dynamics. Instead, we assume that we have access to only a weak prior model of the form $\priordyn(\st,\at,\state_{t+1})$, and estimate a local linear model of the dynamics under this prior. The local linear model is updated at each time step, which allows our method to gradually adapt to changing dynamics conditions and compensate for inaccuracies in the prior model.

\section{Model-Based Reinforcement Learning with Online Dynamics Adaptation}

Our model-based reinforcement learning approach with online dynamics adaptation consists of using MPC (with iterative LQR) to repeatedly update the robot's policy under an evolving dynamics model. The dynamics model is linear time-varying, but is recomputed at each time step based on the recently observed states and actions, as well as the dynamics prior, which can take on one of a number of different representations and is trained on previous robot experience, which might even be from a different task. In this section, we describe how linear dynamics can be fitted under a dynamics prior, as well as our scheme for updating the dynamics online based on the robot's recent experience. We then summarize our model-based reinforcement learning algorithm.

\subsection{Fitting Dynamics with Priors}

In order to fit a linear model of the dynamics to a set of $\empn$ samples $\{\state_i,\action_i,\state_i\pr\}$, we can simply use standard linear regression to determine $\fx$, $\fu$, and $\fc$. To make it more convenient to incorporate prior information, we will first reformulate this linear regression fit and view it as fitting a Gaussian model to the dataset $\{\state,\action,\state\pr\}$, and then conditioning this Gaussian to obtain $p(\state_{t+1}|\st,\at)$. While this is equivalent to linear regression, it allows us to easily incorporate a normal inverse-Wishart prior on this Gaussian in order to bring in prior information. Let $\empsig$ be the empirical covariance of our dataset, and let $\empmu$ be the empirical mean. The normal-inverse-Wishart prior is defined by prior parameters $\priorphi$, $\priormu$, $\priorm$, and $\priorn$. Under this prior, the maximum a posteriori estimates for the covariance $\Sigma$ and mean $\mu$ are given by
\begin{align}
\label{eqn:priorupdate}
\Sigma &= \frac{\priorphi + \empn\empsig + \frac{\empn\priorm}{\empn + \priorm}(\empmu - \priormu)(\empmu - \priormu)\tr}{\empn + \priorn} \nonumber \\
\mu &= \frac{\priorm \priormu + \priorn \empmu}{\priorm + \priorn}.
\end{align}
Having obtained $\Sigma$ and $\mu$, we can obtain an estimate of the dynamics $p(\state_{t+1}|\st,\at)$ by conditioning the distribution $\gauss(\mu,\Sigma)$ on $[\st;\at]$, which produces linear-Gaussian dynamics $p(\state_{t+1}|\st,\at) = \gauss(\fxt\st + \fut\at + \fct, \noise_t)$, given by
\begin{align}
\label{eqn:dyncondition}
\fxu &= \Sigma_{[\xu,\xu]}^{-1}\Sigma_{[\xu,\state\pr]} \nonumber \\
\fc &= \mu_{[\state\pr]} - \fxu\mu_{[\xu]} \nonumber \\
\noise &= \Sigma_{[\state\pr,\state\pr]} - \fxu\Sigma_{[\xu,\xu]}\fxu^T
\end{align}
where the bracketed subscripts denote the submatrix corresponding to the specified elements. The parameters of the normal-inverse-Wishart prior are obtained from some prior model of the dynamics. We discuss how a variety of global dynamics model, such as Gaussian mixture models and neural networks, can be used to produce the prior parameters $\priorphi$, $\priormu$, $\priorm$, and $\priorn$ in Section~\ref{sec:priors}.


\subsection{Online Estimation of Locally Linear Dynamics}
\label{sec:online}

In the batch setting, such as the one described in prior work \cite{la-lnnpg-14}, the empirical mean $\empmu$ and covariance $\empsig$ are estimated from a batch of previously collected data. In this work, we instead would like to update the model online during execution. To do this efficiently, we use an online estimate for $\empmu$ and $\empsig$. The mean can be estimated simply by using the following update:
\begin{equation}
\empmu_t \leftarrow \onlinediscount \empmu_{t-1} + (1 - \onlinediscount) \datapt_t,\label{eqn:empmu}
\end{equation}
\noindent where $\datapt_t = [\state_{t-1} ; \action_{t-1} ; \state_t]$ is the $t^\nth$ observation and $\onlinediscount$ is a discounting factor that causes the model to forget old data. For the covariance, first observe that, in the batch case, $\empsig = \frac{1}{N}\sum_{i=1}^N \datapt_i \datapt_i\tr - \empmu \empmu\tr$. We can therefore estimate $\empsig_t$ as $\empsig_t = \xxt_t - \empmu_t\empmu_t\tr$, where $\xxt_t$ is the current estimate of $\frac{1}{t}\sum_{t\pr=1}^t \datapt_{t\pr} \datapt_{t\pr}\tr$, which we update according to:
\begin{equation}
\xxt_t \leftarrow \onlinediscount \xxt_{t-1} + (1 - \onlinediscount) \datapt_t \datapt_t\tr.\label{eqn:empxxt}
\end{equation}
These updates allow us to estimate the empirical mean $\empmu_t$ and covariance $\empsig_t$ at the $t^\nth$ step using recently observed state transitions. The posterior mean and covariance $\Sigma$ and $\mu$ can then be recovered using the prior and the method described in the previous section. We initialize $\empmu_0$ and $\xxt_0$ by fitting a single Gaussian to the data used for training the prior, in order to start with a reasonable estimate of the dynamics.

The value of $\onlinediscount$ determines how quickly the algorithm ``forgets'' past experiences. When the true dynamics are nonlinear, as in most robotic manipulation tasks, we should forget past experiences more quickly when the state is changing rapidly. However, when the robot becomes stuck in a difficult situation that requires a more accurate dynamics estimate, it must incorporate more data, which requires a larger value of $\onlinediscount$. The effective sample size $\empn$ is also important, since it controls the relative strength of the prior. The true effective sample size is in fact inversely proportional to $1 - \onlinediscount$. We adaptively adjust both $\onlinediscount$ and $\empn$ based on the relative accuracy of the empirical and prior dynamics estimates. The intuition is that, when the prior is less effective, we should weight the empirical estimate more highly. We should also raise $\onlinediscount$, since the empirical estimate factors more highly in the dynamics, and therefore must use more of the past data to improve its accuracy. This intuition is also supported by the inverse relationship between $1 - \onlinediscount$ and $\empn$.

Specifically, we can form the prediction for the most recently observed transition from $(\state_{t-1},\action_{t-1})$ to $\st$ using both the empirically estimated parameters $\empsig_t$ and $\empmu_t$ and the prior parameters $\frac{1}{\priorn}\priorphi_t$ and $\priormut$. We condition both $\gauss(\empmu_t,\empsig_t)$ and $\gauss(\priormut,\frac{1}{\priorn}\priorphi_t)$ on $(\state_{t-1},\action_{t-1})$ and predict the most probable value for the current state, which we denote $\hat{\state}_t$ for the prediction from the empirical parameters and $\bar{\state}_t$ for the prediction from the prior parameters. We then compute the ratio of the errors in these predictions, to compare whether the prior or empirical estimate is more accurate and update $\onlinediscount$ and $\empn$:
\begin{equation}
\rho = \frac{\vnorm{\hat{\state}_t - \st}^2}{\vnorm{\bar{\state}_t - \st}^2} \hspace{0.25in} \onlinediscount = 1 - \eta_0 \rho \hspace{0.25in} \empn = \nu_0 / \rho, \label{eqn:gammaupdate}
\end{equation}
\noindent where $\eta_0$ and $\nu_0$ are hyperparameters of the algorithm. Note that, in these updates, $\empn$ is inversely proportional to $1 - \onlinediscount$, as expected, and the proportionality constant is controlled by $\eta_0$ and $\nu_0$, which we set as $\nu_0 = 1$ and $\eta_0 = 8$ in all of our experiments.

\subsection{Algorithm Summary}

\begin{algorithm}[t]
\caption{Model-based reinforcement learning with online adaptation}
\label{alg:mpc}
\begin{algorithmic}[1]
\FOR{time step $t=1$ to $T$}
\STATE Observe state $\st$
\STATE Update $\empmu_t$ and $\xxt_t$ via Equations~(\ref{eqn:empmu}) and (\ref{eqn:empxxt})
\STATE Compute $\empsig_t = \xxt_t - \empmu_t\empmu_t\tr$
\STATE Evaluate prior to obtain $\priorphi$, $\priormu$, $\priorm$, and $\priorn$ (see Section~\ref{sec:priors})
\STATE Update $\onlinediscount$ and $\empn$ as described in Equation~(\ref{eqn:gammaupdate})
\STATE Compute $\mu$ and $\Sigma$ via Equation~(\ref{eqn:priorupdate})
\STATE Compute $\fxt$, $\fut$, $\fct$, and $\noise_t$ from $\mu$ and $\Sigma$ via Equation~(\ref{eqn:dyncondition})
\STATE Run LQR to compute $\Kpol_t$, $\kpol_t$, and $\Quut$
\STATE Sample $\at$ from $\gauss(\ath + \kpol_t + \Kpol_t (\st - \sth), \Quut\inv)$
\STATE Take action $\at$
\ENDFOR
\end{algorithmic}
\end{algorithm}

The structure of our method is summarized in Algorithm~\ref{alg:mpc}. At each time step, the method observes the current state $\st$ and uses $[\state_{t-1},\action_{t-1},\st]$ to update the current estimate of the empirical mean and covariance $\empmu$ and $\empsig$ as described in the previous section. The method then evaluates the prior to obtain $\priorphi$, $\priormu$, $\priorm$, and $\priorn$, and updates $\onlinediscount$ and $\empn$ based on the accuracy of the empirical and prior state prediction. The empirical and prior estimates are then combined according to Equation~(\ref{eqn:priorupdate}) to construct the posterior mean and covariance, from which we can obtain an estimate for the current dynamics. These estimated dynamics are then used, together with a local second order expansion of the cost function, to optimize a new linear feedback policy using LQR. This feedback policy, given by \mbox{$\detpolicy(\state_t) = \haction_t + \kpol_t + \Kpol_t(\state_t - \hstate_t)$}, can then be used to choose the next action $\at$.

In practice we often want to perform a small amount of exploration. For example, if the robot is attempting a peg insertion task, and the peg becomes jammed in the hole, simply applying the estimated optimal action repeatedly may be insufficient. The robot must ``wiggle'' its arm to figure out the contact dynamics. To that end, we add a small amount of Gaussian noise to the action. The amount of noise to add is determined by the $Q$-function obtained from LQR, by setting the covariance to be proportional to $\Quut\inv$. This choice of covariance is motivated by the observation that it produces a maximum entropy policy that properly trades off randomness and cost minimization, as discussed in prior work \cite{lk-gps-13}.

\section{Neural Network Dynamics Priors}
\label{sec:priors}

In this section, we discuss several possible choices for the dynamics prior, describe how the normal-inverse-Wishart prior parameters can be obtained from these priors, and go into detail on the particular neural network prior that we use in this work. The various choices for the priors are compared in our experimental evaluation in Section~\ref{sec:results}.

\subsection{Gaussian and Gaussian Mixture Priors}

As discussed in the previous section, our online estimate of the dynamics linearization makes use of a dynamics prior. The simplest prior can be obtained by fitting a Gaussian distribution to vectors $[\state; \action; \state\pr]$ obtained from a large batch of prior data. If the mean and covariance of this prior data are given by $\bar{\mu}$ and $\bar{\Sigma}$, the prior is given by $\priorphi = \priorn\bar{\Sigma}$ and $\priormu = \bar{\mu}$, while $\priorn$ and $\priorm$ should be set to the number of data points in the prior datasets. In practice, setting $\priorn$ and $\priorm$ to $1$ tends to produce substantially better results, since the online empirical mean and covariance are typically obtained from a much smaller number of samples.

A more sophisticated choice of prior explored in previous work is a Gaussian mixture model over vectors $[\state; \action; \state\pr]$, which allows modeling of nonlinear dynamics \cite{la-lnnpg-14}. Under this model, the state transition tuple is assumed to come from a distribution that depends on some hidden state $h_i$, which corresponds to the mixture element identity. In practice, this hidden state might correspond to the type of contact profile experienced by a robotic arm at step $i$. The prior is obtained by inferring the hidden state $h_i$ for the latest tuple $[\state_{i-1}; \action_{i-1}; \state_i]$, and using the mean and covariance of the corresponding mixture element to obtain $\bar{\mu}$ and $\bar{\Sigma}$. The prior parameters can then be obtained as described above.

\subsection{Neural Network Priors}

The Gaussian prior is limited to representing globally linear dynamics, while the Gaussian mixture model can only represent a small number of locally linear modes, and has limited representational capacity in complex state spaces. Recent work has shown that neural networks can learn very good dynamics models for tasks ranging from helicopter flight \cite{pa-dlhdm-15} to cutting vegetables \cite{lks-dmpc-15}. We therefore also evaluated the use of neural networks for representing the dynamics prior, by training a neural network to map from state-action tuples $[\state; \action]$ to the next state $\state\pr$.

In order to construct a dynamics prior from a neural network $\net([\state; \action])$ at the current state and action $[\state_i;\action_i]$, we first linearize to obtain
\[
\net([\state;\action]) \approx \net([\state_i;\action_i]) + \frac{d \net}{d [\state; \action]}\tr \left([\state;\action] - [\state_i;\action_i]\right).
\]
From this linearization, we can construct a local prior mean and covariance according to
\begin{align*}
\bar{\mu} &= \left[\begin{matrix}[\state_i;\action_i] \\ \net([\state_i;\action_i])\end{matrix}\right] \\
\bar{\Sigma} &= \left[\begin{matrix}\sigt &  \frac{d \net}{d [\state; \action]}\tr \sigt \\ \sigt \frac{d \net}{d [\state; \action]} & \frac{d \net}{d [\state; \action]}\tr \sigt \frac{d \net}{d [\state; \action]} + \sigp \end{matrix}\right].
\end{align*}
The prior state-action covariance $\sigt$ determines the strength of the prior, and we set it to $\alpha \mathbf{I}$, where $\alpha$ is a free parameter. The conditional covariance $\sigp$ can be obtained from the empirical covariance between the network predictions $\net([\state; \action])$ and the target next states $\state\pr$ in the training dataset. Using the above $\bar{\mu}$ and $\bar{\Sigma}$, the normal-inverse-Wishart prior parameters can be obtained in the same way for the Gaussian and mixture of Gaussians priors.

\subsection{Neural Network Architectures}
\label{sec:arch}

\begin{figure}
\setlength{\unitlength}{0.5\columnwidth}
\begin{picture}(1.99,1.68) \linethickness{0.5pt}
    \includegraphics[width=230pt]{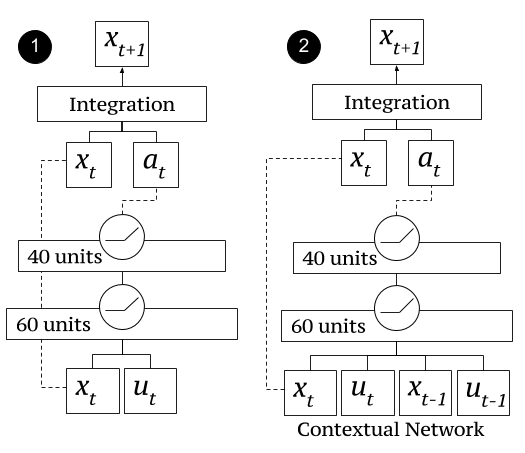}

\end{picture}
\caption{Diagram of the neural network architectures used in our experiments. We found that using a short temporal context as input, as shown in network (2), improved the results for manipulation tasks that involved contact dynamics. Both networks produce accelerations which are used to predict the next state.
\label{fig:arch}
}
\vspace{-0.2in}
\end{figure}

In order to represent the dynamics prior, we used a neural networks with two hidden layers, each with rectified linear units given by $z = \max(a,0)$. The first hidden layer consists of $60$ hidden units, and the second consists of $40$. We evaluated two network architectures, both of which are illustrated in Figure~\ref{fig:arch}. The first takes the current state $\st$ and current action $\at$ as input, and predicts accelerations. These accelerations are then integrated via a semi-implicit integration rule to obtain a prediction for $\state_{t+1}$, and the network is optimized to minimize the error in predicting the entire next state $\state_{t+1}$.

Our second architecture is identical to the first, but instead takes as input the current and previous states and actions $(\state_{t-1},\action_{t-1},\st,\at)$. This additional temporal context is very important when the dynamics prior is trained on other tasks, especially tasks that involve contacts with objects in the environment. Since the state $\st$ only includes the configuration of the robot, it does not model the geometry of the scene, which might change across tasks. The state is therefore not Markovian across tasks, and a network without additional context cannot accurately predict the next state $\state_{t+1}$ from only the current state $\st$ and action $\at$. We found that the network without temporal context tried to explain contacts by exploiting certain regularities. For example, when the end-effector stopped after hitting an object, the network assumed that a hard contact had occurred and there is an impassable obstruction in the way. With a temporal context that included the previous state and action, the network was able to determine whether or not a contact was happening by comparing the previous applied joint torques to the acceleration actually experienced by the robot.

\subsection{Prior Training Data}

The neural network prior must be trained on previous interaction data in order to provide a helpful prior model of the system dynamics. However, because the neural network only acts as a prior, it can be trained on previous interaction data from different tasks. This allows our method to perform one-shot learning of new manipulation tasks, using no prior data from that task itself. In practice, the amount of data required to learn an effective prior is considerable, so we used data collected from a variety of sources to provide a sufficiently diverse dataset. The total training set for the physical robot experiments had 6.6 hours of data, collected at 20 Hz. For each individual task, we excluded the data collected for that task when training the prior, which reduced the effective dataset by about 15\%. This corresponds to a type of holdout cross-validation. Likewise, in the simulated experiments, we collected approximately 12 simulation hours of data at 20Hz across 4 different tasks, and excluded the data from the task being executed from the training set.

\begin{figure}
\setlength{\unitlength}{0.5\columnwidth}
\begin{picture}(1.99,1.09) \linethickness{0.5pt}
    \includegraphics[height=125pt]{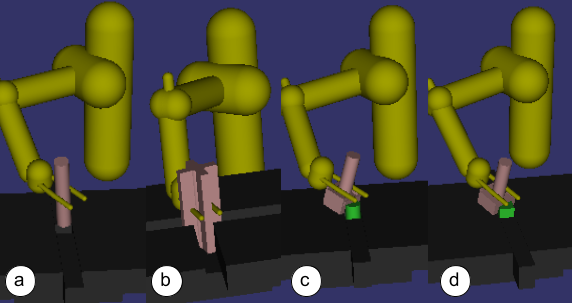}
\end{picture}
\caption{ Simulated tasks used for evaluation: (a) Cylindrical peg insertion
(b) Cross-shaped peg insertion
(c) Stacking over a cylindrical peg
(d) Stacking over a square-shaped peg
\label{fig:sim_tasks}
}
\vspace{-0.25in}
\end{figure}

The dataset consisted of trials from the various evaluation tasks (workbench, gears, airplane, car, and ring tasks for the physical system and peg insertion and stacking for the simulated system), which are described in the next section, as well as random motion of the arm in free space. For each task, data was collected using a previous reinforcement learning method~\cite{lwa-lnnpg-15}, though in practice this data could have also been generated using prior experience from the method presented in this paper, human demonstrations, or any other method that provides good coverage of interesting states (e.g. states that involve contact with objects in the world). Although the size of this dataset is considerable, the resulting prior, when combined with our online adaptation method, can generalize effectively to other tasks, making it quite universal.





\section{Experimental Evaluation}

\label{sec:results}

\begin{figure*}
\setlength{\unitlength}{0.5\columnwidth}
\begin{picture}(1.99,1.09) \linethickness{0.5pt}
    \includegraphics[height=125pt]{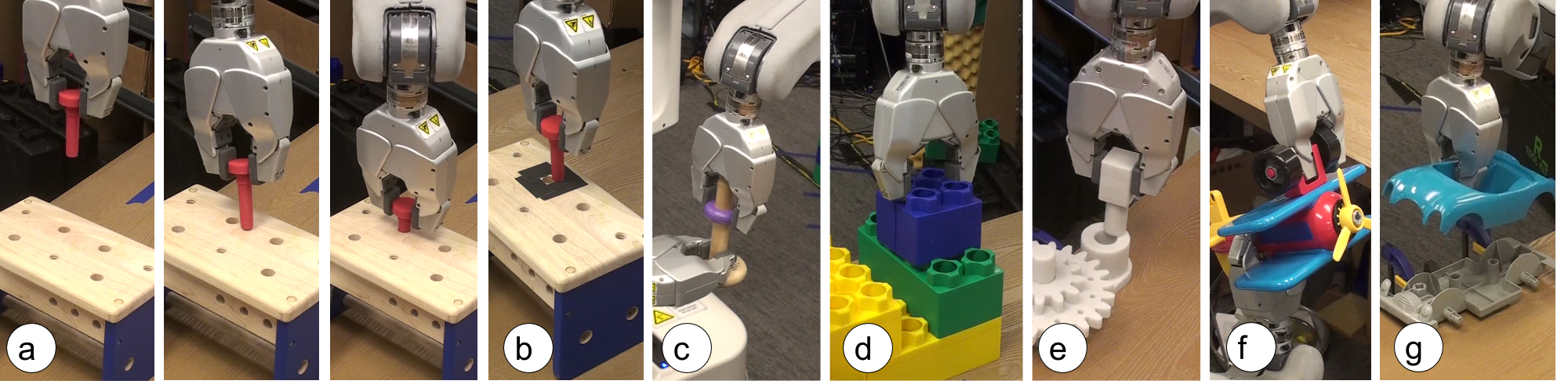}

\end{picture}
\caption{Tasks used in our evaluation: 
(a) inserting a toy nail into a toolbench,
(b) inserting the nail with a high-friction surface to increase difficulty,
(c) placing a wooden ring on a tight-fitting peg,
(d) stacking toy blocks,
(e) putting together part of a gear assembly,
(f) assembling a toy airplane and
(g) a toy car.
\label{fig:tasks}
}
\vspace{-0.1in}
\end{figure*}

\begin{table*}[t]
\parbox{.99\textwidth}{
\begin{tabular}{| l || l | l | l | l | l | l | l | l | l | l | l | l | l }
\hline
Task & \multicolumn{1}{c|}{Ins, Cylinder (a)} & \multicolumn{1}{c|}{Ins, Cross (b)} & \multicolumn{1}{c|}{Stack, Cylinder (c)} & 
\multicolumn{1}{c|}{Stack, Square (d)} &
Total \\
\hline
Adaptation + regularization & 10/25 & 5/25 & 18/25 & 10/25 & 43/100\\
\hline
Adaptation + GMM prior & 10/25 & 6/25 & \bf 25/25 &18/25 &  59/100\\
\hline
Contextual network (\#2) & 8/25 & 1/25 & 19/25 &12/25 &  40/100 \\
\hline
Adaptation + network \#1 & 14/25 & 8/25 & 20/25 &14/25 &  56/100 \\
\hline
Adaptation + Contextual network (\#2) ({\bf full method}) & \bf 19/25 & \bf 11/25 & \bf 25/25 & \bf 20/25 &  \bf 75/100 \\
\hline
\end{tabular}
\caption{Success rate of our method on simulated tasks, as well as comparisons to a range of baselines representative of prior methods on three of the tasks.
\label{tbl:results_sim}
}
}
\vspace{-0.1in}
\end{table*}

\begin{table*}[t]
\parbox{.70\textwidth}{
\begin{tabular}{| l || l | l | l | l | l | l | l | l | l | l | l | l | l |}
\hline
Task & \multicolumn{1}{c|}{Toolbench (a)} & \multicolumn{1}{c|}{W/ friction (b)} & \multicolumn{1}{c|}{Ring (c)} & Total \\
\hline
Adaptation + regularization & 2/5 & 0/5 & \bf 4/5 & 6/15\\
\hline
Adaptation + GMM prior & 0/5 & 1/5 & 2/5 & 3/15\\
\hline
Gaussian process & 0/5 & 0/5 & 0/5 & 0/15\\
\hline
Adaptation + gaussian process & 1/5 & 0/5 & 0/5 & 1/15\\
\hline
Contextual network (\#2) & 3/5 & 3/5 & \bf 4/5 & 10/15 \\
\hline
Adaptation + network \#1 & 4/5 & 0/5 & 3/5 & 7/15 \\
\hline
Adaptation + Contextual network (\#2) ({\bf full method}) & \bf 5/5 & \bf 4/5 & \bf 4/5 & \bf 13/15 \\
\hline
\end{tabular}

\caption{Success rate of our method on each physical task. Note that our approach regularly succeeds at the task despite only using training data from other tasks.
\label{tbl:results}
}
}
\vspace{-1.3in} 
\hfill
\hspace{0.34in}
\parbox{.25\textwidth}{
\begin{tabular}{| l || l |}
\hline
Task & Success Rate \\
\hline
Block (d) & 4/5 \\
\hline
Gears (e)& 4/5 \\
\hline
Airplane (f)& 3/5 \\
\hline
Car (g) & 3/5 \\
\hline
\end{tabular}
\caption{Success rates on additional tasks using adaptation with network \#2 only.}
\label{tbl:results_extra}
}
\vspace{1.05in} 
\end{table*}

\subsection{Experimental Setup}

We evaluated our method on a range of manipulation tasks using a physical PR2 mobile manipulator and a simulated PR2 arm using the MuJoCo simulator \cite{tet-mjc-12}. The simulated tasks provide a large-scale comparison between methods, and the physical tasks demonstrate effectiveness on a real system.

On the physical system, our method was used to control the right arm of the robot, while the left arm was used to brace objects for two-object manipulation and assembly tasks. We evaluated the method by inserting a toy nail into a tool bench, including a high-friction version of the task designed to make it more difficult, placing wooden rings onto pegs, putting together parts of a gear assembly, assembling a toy car and airplane, and stacking blocks. These tasks are shown in Figure~\ref{fig:tasks}. Our simulated tasks consisted of a pair of peg insertion tasks with different shaped pegs, and a pair of peg stacking tasks, where the robot needs to fit a piece with a square-shaped opening over a cylindrical or square shaped peg. These are shown in Figure~\ref{fig:sim_tasks}. A trial was defined to be a failure if the distance to a target pose was not sufficiently small at the end of a time threshold of 10 seconds.

The state space consisted of several measurements from the active right arm of the robot: joint angles, joint angular velocities, the pose of the end effector encoded as 3 Cartesian points, and the velocities of those 3 points. This amounted to a 32-dimensional state space. The controller operated at 20 Hz with an MPC horizon of 15 timesteps (0.75 seconds).

The cost function for each task depended on the distance $d$ between the current pose of the end-effector and a target pose required for successful completion. For example, in the case of the toy nail task, the target pose involved positioning the gripper such that the nail was successfully inserted into a hole in the toy tool bench. The shape of this cost follows prior work \cite{lwa-lnnpg-15}, and is given by $\costnorm(d) = w d^2 + v \log(d^2 + \alpha)$, where $\alpha = 10^{-5}$, $w = 1.0$, and $v = 0.01$. This type of shape encourages the robot to accurately place the object in precisely the desired configuration. In addition, we placed a quadratic penalty on the magnitude of the torques.

As described in the previous section, the neural network dynamics prior was trained using data collected from all of the other tasks, but no data from the task being tested. This corresponds to a hold-one-out cross-validation scheme.

\subsection{Comparisons}

We evaluated our adaptive method with two neural network priors: network \#1 in Figure~\ref{fig:arch}, which uses only the current state $\st$ and action $\at$ to predict the next state $\state_{t+1}$, and network \#2, which also uses the preceding state and action. In addition, we also compared to a number of baselines. The non-adaptive baseline used network \#2, which was the better of the two neural network priors, to directly plan actions using MPC, which most closely resembles the structure of a previously proposed neural network-based MPC method \cite{lks-dmpc-15}. 
The Gaussian process baselines used a Gaussian process either as a prior for the adaptive approach, or as the model (without adaption) for MPC, which most closely reflects a range of recent Gaussian process model-based reinforcement learning algorithms \cite{dr-pmbde-11,pt-pddp-14}. Since the Gaussian process is a nonparametric model, we were unable to use the same amount of data with this model and maintain real time performance. We therefore subsampled the data to the largest size that still permitted online MPC (approximately 10000 training points, which was about 5\% of the total training set for the physical robot). Finally, the regularized model substituted a global linear model instead of the neural network dynamics prior.

We show the success rates for each method on the simulated robot in Table~\ref{tbl:results_sim} and on the physical robot in Tables~\ref{tbl:results} (for the standard and high-friction variant of the toolbench and toy nail task, as well as for the wooden ring on a peg task) and~\ref{tbl:results_extra} (for the block, gears, airplane, and car assembly tasks). Note that each of the runs used to evaluate each method was performed separately, with no information retained between runs. The adaptive method outperformed the non-adaptive variants, indicating the importance of online adaptation for models trained on other tasks. The neural network prior also achieved the best overall performance, particularly when using the preceding state and action as context, although even the simple least-squares prior was able to accomplish some of the simpler tasks.

We also show the success rates for our adaptive method with network \#2 on each of the other tasks. These results show that our method was able to succeed on a wide range of challenging manipulation tasks on the first try, without using any data from that task to train the dynamics prior.

\subsection{Robustness}

\begin{table*}[t]
\vspace{0.15in} 
\begin{center}
\begin{tabular}{| l || l | l | l || l | l | l || l | l | l |}
\hline
Task & \multicolumn{3}{c||}{Block} & \multicolumn{3}{c|}{Toolbench} & \multicolumn{3}{c|}{Ring} \\
\hline
Target position error & 0.5 cm & 1.0 cm & 1.5 cm & 0.5 cm & 1.0 cm & 1.5 cm & 0.5 cm & 1.0 cm & 1.5 cm \\
\hline
Contextual network (\#2) & 2/5 & 0/5 & 0/5 & 4/5 & 2/5 & 0/5 & 4/5 & 4/5 & 1/5 \\
\hline
Adaptation + contextual network (\#2) ({\bf full method}) & 4/5 & 1/5 & 0/5 & 3/5 & 3/5 & 1/5 & 4/5 & 3/5 & 0/5 \\
\hline
\end{tabular}
\end{center}
\vspace{-0.1in}
\caption{Robustness results in the presence of target position errors. The target was intentionally offset from the true position of the target by the indicated amount in random direction along the horizontal plane.
\label{tbl:robustness}
}
\vspace{-0.2in}
\end{table*}

We also evaluated the robustness of our method to observation errors. In these experiments, the target position in the cost function was intentionally corrupted by fixed magnitude errors. In real-world scenarios, these errors might stem from imperfect observations produced, for example, by a vision system. The results of the robustness experiments on two of the tasks are presented in Table~\ref{tbl:robustness}. For comparison, we also include robustness results for the neural network only (using network \#2), without adaptation. The ring was the easiest of the three tasks, since the rounded top of the peg provides some tolerance, while the block was the hardest. Note that adaptation is particularly helpful on the harder tasks.

\subsection{Qualitative Results and Conclusions}

Our experimental results show that our model-based reinforcement learning algorithm with online adaptation can achieve a range of challenging manipulation tasks on the first attempt. Furthermore, our comparison of the various prior models shows that a neural network prior with a short temporal context achieved the best results, though no prior model by itself was as successful as the corresponding adaptive variant. Videos of our tasks, which can be viewed in the supplementary material and on the project website, (\url{http://rll.berkeley.edu/iros2016onlinecontrol/index.html}) show that the resulting behaviors can take some time to succeed at the task (though never more than 20 seconds). Much of this time is spent exploring unfamiliar dynamical modes, for example after an unexpected contact. This kind of exploration reflects a very natural strategy for performing unfamiliar tasks: probing the object until progress is made in the desired direction, while gradually acquiring a more accurate model. It is precisely this probing and gradual self-improvement that is key to the robustness and effectiveness of this approach.

\subsection*{Acknowledgements}
This research was funded in part by the Army Research Office through the MAST program, and by Darpa under Award \#N66001-15-2-4047.

\section{Discussion and Future Work}

In this paper, we presented a model-based reinforcement learning algorithm that uses online adaptation of its dynamics model combined with a coarse prior model obtained from experience on other prior tasks. We use an MPC method based on iterative LQR to choose the actions under the current adapted model, which is a local linear approximation to the nonlinear dynamics of the system. Although this linear model is simple, our method is able to perform complex manipulation tasks in highly nonlinear systems by adapting it to the most recent experience and re-planning. The coarse prior model is therefore not required to accurately model the true dynamics of the system, and only provide a guess that is refined through online adaptation. Our experiments show that even a fully linear prior model can be used for some behaviors, while a more sophisticated neural network prior allows our method to complete more complex tasks.

We believe that combining online model adaptation with prior experience by means of a dynamics prior is a powerful idea because it allows our method to succeed even with unexpected environmental variation and inaccurate prior models. Furthermore, by using rich function approximators such as large neural networks to distill prior experience into a concise parametric representation, the robot can progressively become more proficient at acquiring new skills as its experience grows, similar to how a person can quickly learn new skills by drawing on past experiences. However, in contrast to humans, robots can pool their collective experience and use the combined data to train shared dynamics priors. Exploring this type of multi-robot learning, with shared priors but individual adaptation, is an exciting direction for future work.

While in our evaluations, we aggregated data from all other tasks to train the dynamics model, we may want to only use related tasks to avoid negative transfer. This may be accomplished by clustering or grouping the prior tasks.

Although we demonstrate that our approach can complete a variety of challenging manipulation tasks, a limitation of this method is that we require a full, Markovian state of the system, which is needed to perform MPC using iterative LQR. This assumption is not unusual for MPC-based methods, but can be limiting in the context of robotic manipulation. One approach for addressing this is to learn a latent state representation from raw sensory input, as proposed in recent work \cite{rlv-arlrv-12,wpbr-etcll-15,finn2016autoenc}, and perform MPC on this learned representation with online dynamics adaptation.

Another avenue for future work is to combine other kinds of prior information with online updates. For example, a prior policy might be constructed from experience of related tasks, and refined in a similar fashion as the dynamics adaptation. This could allow our relatively short-horizon MPC procedure to acquire longer lookahead through the use of the prior policy, and allow the use of rich sensory data as demonstrated in recent work on policy search with neural networks \cite{levine2015end}.



\bibliographystyle{IEEEtran}
\bibliography{references}

\end{document}